# Deep MR Fingerprinting with total-variation and low-rank subspace priors

Mohammad Golbabaee[1], Carolin M. Pirkl[2,4], Marion I. Menzel[4], Guido Buonincontri, and Pedro A. Gómez[3,4]

[1]Computer Science Department, University of Bath, [2]Computer Science Department, Technische Universität München, [3]School of Bioengineering, Technische Universität München, [5]GE Healthcare, Munich

**Synopsis:**

Deep learning (DL) has recently emerged to address the heavy storage and computation requirements of the baseline dictionary-matching (DM) for Magnetic Resonance Fingerprinting (MRF) reconstruction. Fed with non-iterated back-projected images, the network is unable to fully resolve spatially-correlated corruptions caused from the undersampling artefacts. We propose an accelerated iterative reconstruction to minimize these artefacts before feeding into the network. This is done through a convex regularization that jointly promotes spatio-temporal regularities of the MRF time-series. Except for training, the rest of the parameter estimation pipeline is dictionary-free. We validate the proposed approach on synthetic and in-vivo datasets.

**Purpose:**

Dictionary-matching (DM) approaches proposed for MR Fingerprinting (MRF)[1-4] do not scale well to the growing complexity of the emerging multi-parametric quantitative MRI problems[5,6]. Deep learning (DL) methodologies are recently introduced to overcome this problem[7-9]. Time-series of Back-Projected Images (BPI) are fed into a compact neural network which temporally processes voxel sequences and approximates the DM step to output the parametric maps. Trained by independently corrupted (i.i.d. Gaussian) noisy fingerprints, the network is unable to correct for dominant spatially-correlated (aliasing) artefacts appearing in BPIs in highly undersampled regimes. Also larger DL models aiming to learn spatio-temporal data structures[10,11] are prone to overfitting due to the limited access to properly large ground-truth parametric maps (i.e. spatial joint distributions of the quantitative MR parameters) in practice. Further, such approaches build customized de-noisers which require expensive re-training by changing sampling parameters i.e. the forward model.

This abstract aims to address these shortcomings by taking a dictionary-free compressed sensing approach to spatio-temporally process data before feeding into a compact and easily-trained network of the first type. Casted in a convex problem, the spatial regularities of the MRF time-series are promoted by Total Variations[12] (TV) shrinkage and temporal structures are relaxed to low-rank subspace constraints.

**Parameter estimation pipeline:**

The undersampled k-space measurements $Y \in \mathbb{C}^{m \times L}$ acquired across $L$ timeframes are first processed by solving the following convex and dictionary-free regularized inverse problem:

$$\hat{X} = \mathrm{argmin}_{X_S} |Y - A(XV_s^H)|_2^2 + \lambda \sum_{i=1}^{S} |X_i|_{TV} \quad (P1)$$

in order to find $S \ll L$ principal/subspace images $X \in \mathbb{C}^{n \times S}$. Subspace bases $V_s \in \mathbb{C}^{L \times S}$ are the $S$ leading (left) SVD components of a large-size MRF dictionary $D \in \mathbb{C}^{L \times d}$ $L \ll d$ used here for unsupervised training[2,13,14]. The forward operator $A$ models the multi-coil sensitivities and the per-frame subsampled 2D Fourier Transforms.

The subspace model is a convex (in fact linear) relaxed representation of the temporal dictionary responses and when accurate enough, it is computationally advantageous over the full image representation $X^{Full} \approx X^S V_s^H \in \mathbb{C}^{n \times L}$ because it reconstructs smaller objects and promotes temporally low-rank structures. This prior alone is, however, insufficient to obtain artefact-free solutions e.g. when using spiral readouts[15,16]. We additionally use TV regularization by choosing $\lambda > 0$ to promote spatial (piecewise) smoothness across recovered subspace images. Optimization (P1) can be efficiently solved using the iterative shrinkage Algorithm 1 with momentum acceleration and backtracking step-size[17-19].

Parameter maps are estimated using the MRF-Net[9] that is a 3-layer fully-connected network (Figure 2). The MRF dictionary is only used for training and not during parameter recovery. Fed with the iteratively reconstructed images $\hat{X}$, the MRF-Net processes each (normalized) voxel sequence and outputs per-voxel quantitative parameters. The MRF-Net has implicitly 4-layers by including the subspace projection that is incorporated in solving (P1). Three other layers include nonlinearities in order to approximate subspace dictionary matching. Thanks to this dimensionality-reduction, MRF-Net requires far less units and training resources compared to the uncompressed DL approaches[7,8].

**Methods:**

Methods are tested on a numerical brain phantom[20], and a healthy human brain (in-vivo data was acquired on a 3T GE MR750w system, GE Medical Systems, Milwaukee, WI, using 8-channel receive-only head RF coil). Acquisition follows the Steady State Precession (FISP) sequence that jointly encodes T1 and T2 values using sinusoidal flip angle variations[3], fixed TR=10msec, TE=1.908msec, Tinv=18msec, L=1000 repetitions, variable-density spiral sampling, 377 interleaves, 22.5x22.5cm2 FOV, 256x256 spatial resolution, 1.3mm in-plane resolution and 5mm slice thickness).

We use the Extended Phase-Graph formalism[21] and simulate a dictionary of $d = 113'640$ fingerprints for combinations of T1=[100:10:4000], T2 = [20:2:600] msec. Clean fingerprints are used for unsupervised subspace model learning of sufficiently low-rank ($S = 10$ here[14]). Further, fingerprints corrupted by additive white Gaussian noise (data augmentation by factor 100) supervisedly train the dimension-reduced MRF-Net on a standard CPU desktop[9]. We adopt a practical phase-alignment heuristic[15] to de-phase dictionary atoms (for training) and subspace images (the inputs). Complex-valued extensions[8] are possible by stacking the real and imaginary components into a larger real-valued object, however, at the cost of extra training for randomly generated complex phases. We find this treatment unnecessary in our experiments.

**Results and discussion:**

We compare three methods for reconstructing subspace images before fed to the MRF-NET: non-iterative BPI i.e. $\hat{X} := A^H(Y)V_s$, and iterative reconstructions incorporating ii) only the low-rank (LR) subspace prior by solving P1 with $\lambda = 0$, and iii) joint TV and subspace spatio-temporal priors (LRTV) by solving P1 with an experimentally tuned $\lambda = 2e - 5$. Note that the BPIs are the first iteration of the LR. Figures 3 and 4 show reconstructed maps for synthetic and scanner data. Undersampling artefacts are visible in BPI+MRF-Net. The subspace iterations of LR+MRF-Net also admit undesirable solutions with high-frequency artefacts due to the insufficient measurements collected from the k-space corners in spiral readouts (for details see[15,16]). By adding sufficient spatial regularization, the proposed LRTV+MRF-Net outputs artefact-free maps within 8-12 iterations.

**Acknowledgements:**

Authors would like to thank Mike Davies, Dongdong Chen and Ian Marshall from University of Edinburgh for helpful discussions during preparation of this draft. This work is supported by the Scottish Research Partnership in engineering (SRPe) award PECRE 1718/18 and the European Commission FP7-PEOPLE 605162 project BERTI.

**References**

[1]     D. Ma, V. Gulani, N. Seiberlich, K. Liu, J. Sunshine, J. Durek, and M. Griswold, "Magnetic resonance fingerprinting," Nature, vol. 495, no. 7440, pp. 187–192, 2013.

[2]     McGivney, D. F., Pierre, E., Ma, D., Jiang, Y., Saybasili, H., Gulani, V., and Griswold, M. A., "SVD compression for magnetic resonance fingerprinting in the time domain. IEEE transactions on medical imaging, 33(12), 2311-2322, 2014.

[3]     N. Jiang Y, D. Ma, N. Seiberlich, V. Gulani, and M. Griswold,"MR fingerprinting using fast imaging with steady state precession (fisp) with spiral readout," Magnetic resonance in medicine, vol. 74, no. 6, pp. 1621–1631, 2015.

[4]     M. Davies, G. Puy, P. Vandergheynst, and Y. Wiaux, "A compressed sensing framework for magnetic resonance fingerprinting," SIAM Journal on Imaging Sciences, vol. 7, pp. 2623-2656, 2014.

[5]     Wang, C.Y., Coppo, S., Mehta, B.B., Seiberlich, N., Yu, X., and Griswold, M.A., "Magnetic Resonance Fingerprinting with Quadratic RF Phase for Simultaneous Measurement of δf, T1, T2, and T2*," in Proc. Intl. Soc. Mag. Reson. Med., 2017.

[6]     K. L. Wright, Y. Jiang, D. Ma, D. C. Noll, M. A. Griswold, V. Gulani, and L. Hernandez-Garcia, "Estimation of perfusion properties with MR fingerprinting arterial spin labeling," Magnetic resonance imaging, vol. 50, pp. 68–77, 2018.

[7]     O. Cohen, B. Zhu, and M. S. Rosen, "MR fingerprinting deep reconstruction network (DRONE)," Magnetic resonance in medicine, vol. 80, no. 3, pp. 885–894, 2018.

[8]     P. Virtue, X. Y. Stella, and M. Lustig, "Better than real: Complex-valued neural nets for mri fingerprinting," in Image Processing (ICIP), 2017 IEEE International Conference on, IEEE, 2017, pp. 3953–3957.

[9]     M. Golbabaee, D. Chen, P. A. Gómez, M. I. Menzel and M. E. Davies. "Geometry of Deep Learning for Magnetic Resonance Fingerprinting." arXiv preprint arXiv:1809.01749, 2018.

[10]    E. Hoppe, G. Körzdörfer, T. Würfl, J. Wetzl, F. Lugauer, J. Pfeuffer, and A. Maier, "Deep learning for magnetic resonance fingerprinting: A new approach for predicting quantitative parameter values from time series." Studies in health technology and informatics, vol. 243, p. 202, 2017.

[11]    F. Balsiger, A. S. Konar, S. Chikop, V. Chandran, O. Scheidegger, S. Geethanath, and M. Reyes, "Magnetic resonance fingerprinting reconstruction via spatiotemporal convolutional neural networks," in InternationalWorkshop on Machine Learning for Medical Image Reconstruction. Springer, 2018, pp. 39–46.

[12]     L. I. Rudin, S. Osher, and E. Fatemi. "Nonlinear total variation based noise removal algorithms." Physica D: nonlinear phenomena 60.1-4 (1992): 259-268.

[13]    Assländer, J., Cloos, M. A., Knoll, F., Sodickson, D. K., Hennig, J., & Lattanzi, R. (2018). Low rank alternating direction method of multipliers reconstruction for MR fingerprinting." Magnetic resonance in medicine, 79(1), 83-96.

[14]    B. Zhao, K. Setsompop, E. Adalsteinsson, B. Gagoski, H. Ye, D. Ma, Y. Jiang, P. Ellen Grant, M. A. Griswold, and L. L. Wald, "Improved magnetic resonance fingerprinting

reconstruction with low-rank and subspace modeling," Magnetic resonance in medicine, vol. 79, no. 2, pp. 933–942, 2018.

[15]     M. Golbabaee, Z. Chen, Y. Wiaux, and M. Davies, M. "CoverBLIP: accelerated and scalable iterative matched-filtering for Magnetic Resonance Fingerprint reconstruction." arXiv preprint arXiv:1810.01967, 2018.

[16]     Cline, C.C., Chen, X., Mailhe, B., Wang, Q., Pfeuffer, J., Nittka, M., Griswold, M.A., Speier, P. and Nadar, M.S., "AIR-MRF: accelerated iterative reconstruction for magnetic resonance fingerprinting," Magnetic Resonance Imaging, 41, pp.29-40, 2017.

[17]     Y. E. Nesterov. "A method for solving the convex programming problem with convergence rate $O(1/k^2)$," Dokl. Akad. Nauk SSSR, 269 (1983), pp. 543–547 (in Russian).

[18]     A. Beck and M. Teboulle, "A fast iterative shrinkage-thresholding algorithm for linear inverse problems." SIAM journal on imaging sciences 2.1 (2009): 183-202.

[19]     A. Beck and M. Teboulle. "Fast gradient-based algorithms for constrained total variation image denoising and deblurring problems." Image Processing, IEEE Transactions on, 18(11):2419--2434, 2009.

[20]     Brainweb data repository, available at: http://brainweb.bic.mni.mcgill.ca/brainweb/

[21]     Weigel, M., "Extended phase graphs: dephasing, RF pulses, and echoes-pure and simple," *Journal of Magnetic Resonance Imaging,* vol. 41, pp. 266-295, 2015.

[22]     S. Arberet, X. Chen, B. Mailhe, P. Speier, M. S. Nadar. "CS-MRF: sparse and low-rank iterative reconstruction for magnetic resonance fingerprinting", ISMRM workshop on Magnetic Resonance Fingerprinting, Oct, 2017.

**Algorithm 1** for solving P1

1: **Inputs:** k-space data $Y$, forward and adjoint operators $\mathcal{A}, \mathcal{A}^H$,
2: MRF temporal subspace $V_s$, initial step-size $\mu$, regularization parameter $\lambda$.
3: **Initialization:** $k = 1$, $X^1 = \mathbf{0}$, $\mu_k = \mu$ $\forall k = 1, 2, \ldots$
4: **while** stopping criterion = false **do**
5: $\quad \nabla = \mathcal{A}^H \left( \mathcal{A}(X^k V_s^H) \right) V_s - \mathcal{A}^H(Y) V_s$ #(subspace gradient)
6: $\quad Z^k = \mathbf{prox}_{\lambda \mu_k \mathrm{TV}}(X^k - \mu_k \nabla)$ #(total variation shrinkage with threshold $\lambda \mu_k$)
7: $\quad$ **if** $\|Y - A(Z^k V_s^H)\|^2 > \|Y - A(X^k V_s^H)\|^2 + 2\mathrm{Re}\langle \nabla, Z^k - X^k \rangle + \mu_k^{-1} \|Z^k - X^k\|^2$ **then**
8: $\quad\quad \mu_k = \mu_k / 2$ #(adaptive step-size shrinkage)
9: $\quad\quad$ go back to line 6
10: $\quad$ **else**
11: $\quad\quad X^{k+1} = Z^k + \left(\frac{k-1}{k+2}\right)(Z^k - Z^{k-1})$ #(momentum acceleration)
12: $\quad\quad k = k + 1$
**return** reconstructed subspace MRF image $X^{k+1}$

**Figure 1.** Iterative shrinkage algorithm based on[17,18] with momentum acceleration and backtracking step-size for solving P1. Total variation proximal shrinkage (step 6) promoting spatial smoothness is iteratively solved using[19]. Low-rank subspace prior is included in fast dimension-reduced gradient updates thanks to commuting spatio-temporally separable operators $A, V_s$ (see details in[15]). The initial step-size is set to the compression factor $\mu = n/m$ according to[4,15,16]. The convexity, momentum acceleration and dictionary-free (resource efficient) features of Algorithm 1 distinguish it from a related work[22] used spatio-temporal priors with expensive DM iterations.

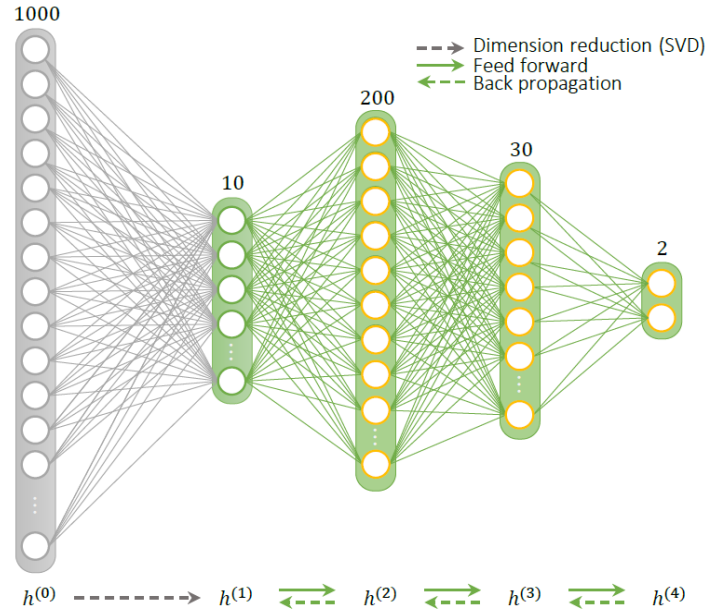

**Figure 2.** Illustration of the MRF-Net proposed in[9]. Inputs $h^{(1)}$ are the voxel sequences of the subspace image reconstructed by Algorithm 1 and outputs $h^{(4)}$ are the per-voxel T1 and T2 parameters. The MRF-Net has implicitly 4 layers by including the unsupervisedly learned subspace projection (first layer in gay) incorporated in solving (P1). Thanks to this dimensionality-reduction, MRF-Net requires less units and training resources compared to the uncompressed DL approaches[7,8]. Three last layers use nonlinear ReLU activations (orange) and are supervisedly trained by standard backpropagation to approximate subspace dictionary matching.

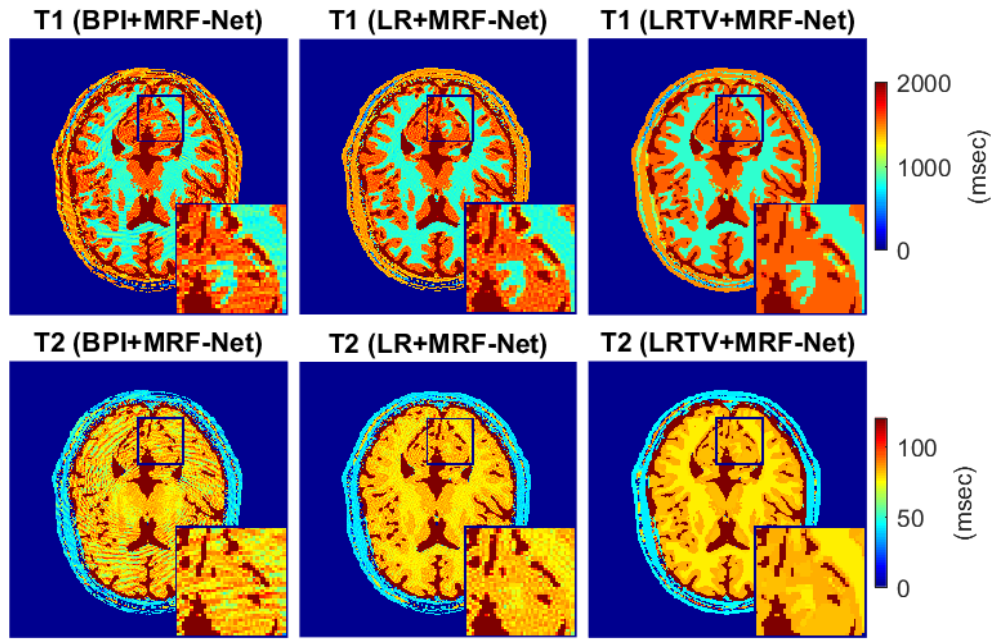

**Figure 3.** Reconstructed T1 (top row) and T2 (bottom row) maps of the numerical Brainweb phantom using MRF-Net fed with the non-iterated BPIs (left column), iteratively reconstructed images with only low-rank (LR) subspace prior (middle column), and iteratively reconstructed images with joint TV and low-rank subspace (LRTV) priors (right column). Undersampling artefacts are visible in BPI+MRF-Net. The subspace iterations of LR+MRF-Net removes them but it admits a solution with high-frequency artefacts due to the insufficient measurements from the corners of the k-space. By adding sufficient spatial regularization, the proposed LRTV+MRF-Net outputs artefact-free maps.

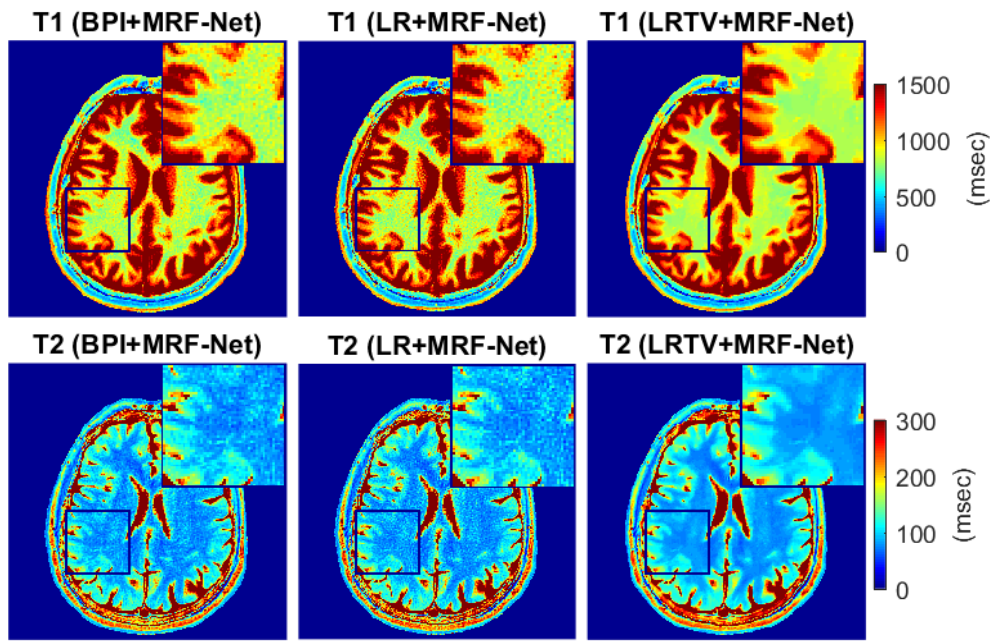

**Figure 4.** Reconstructed T1 (top row) and T2 (bottom row) maps for healthy volunteer data using BPI+MRF-Net (left column), LR+MRF-Net (middle column) and LRTV+MRF-Net (right column) approaches. Multi-coil data helps reducing artefacts using the first two approaches compared to the (simulated) single-coil numerical phantom experiment in Figure 3. However to output clean parameter maps one requires to incorporate jointly the spatial (piece-wise smoothness) and temporal (low-rank subspace) priors through the proposed LRTV+MRF-Net.